\documentclass[a4paper,11pt]{article}

\usepackage{fourier}
\usepackage{color}
 \usepackage{graphicx}
\usepackage{url}
\usepackage[affil-it]{authblk}
\usepackage{algorithm2e}
\usepackage{algpseudocode}
\usepackage{amsmath}
\usepackage{wrapfig}
\usepackage{float}
\usepackage{hyperref}
\usepackage{titlesec}
\usepackage[T1]{fontenc}
\usepackage{times}

\begin{document}
\vspace{-50pt}
\title{KinePose: A temporally optimized inverse kinematics technique for 6DOF human pose estimation with biomechanical constraints}

\author[1]{Kevin Gildea}
\author[1]{Clara Mercadal-Baudart}
\author[1,2]{Richard Blythman}
\author[2]{Aljosa Smolic}
\author[1]{Ciaran Simms}
\affil[1]{School of Engineering, Trinity College Dublin}
\affil[2]{V-SENSE, School of Computer Science \& Statistics, Trinity College Dublin}

%\author{Anonymous Submission}
%\affil{Anonymous Affiliation}
%\vspace{-50pt}
\vspace{10pt}
\date{}
\maketitle
\thispagestyle{empty}

\vspace{-10pt}
\begin{abstract}
Computer vision/deep learning-based 3D human pose estimation methods aim to localize human joints from images and videos. Pose representation is normally limited to 3D joint positional/translational degrees of freedom (3DOFs), however, a further three rotational DOFs (6DOFs) are required for many potential biomechanical applications. Positional DOFs are insufficient to analytically solve for joint rotational DOFs in a 3D human skeletal model. Therefore, we propose a temporal inverse kinematics (IK) optimization technique to infer joint orientations throughout a biomechanically informed, and subject-specific kinematic chain. For this, we prescribe link directions from a position-based 3D pose estimate. Sequential least squares quadratic programming is used to solve a minimization problem that involves both frame-based pose terms, and a temporal term. The solution space is constrained using joint DOFs, and ranges of motion (ROMs). We generate 3D pose motion sequences to assess the IK approach both for general accuracy, and accuracy in boundary cases.
Our temporal algorithm achieves 6DOF pose estimates with low Mean Per Joint Angular Separation (MPJAS) errors (3.7\textdegree/joint overall, \&  1.6\textdegree/joint for lower limbs). With frame-by-frame IK we obtain low errors in the case of bent elbows and knees, however, motion sequences with phases of extended/straight limbs results in ambiguity in twist angle. With temporal IK, we reduce ambiguity for these poses, resulting in lower average errors. Code and supplementary material are available\footnote{\href{https://kevgildea.github.io/KinePose/}{https://kevgildea.github.io/KinePose/\label{p}}}.

\end{abstract}

\vspace{-10pt}

%%%%%%%%%%%%%%%%%%%%%%
\section{Introduction}
%\begin{wrapfigure}{r}{0.5\textwidth}
%  \vspace{-20pt}
%  \begin{center}
%    \includegraphics[width=0.48\textwidth, %height=0.28\textwidth]{CampanileEdges}
%\end{center}
%\vspace{-20pt}
%  \caption{Introduction Banner.}
%  \vspace{-0pt}
%\end{wrapfigure}
%%%%%%%%%%%%%%%%%%%%%%
\vspace{-10pt}
\begin{wrapfigure}{r}{0.6\textwidth}
  \vspace{-25pt}
  \begin{center}
    \includegraphics[width=0.6\textwidth]{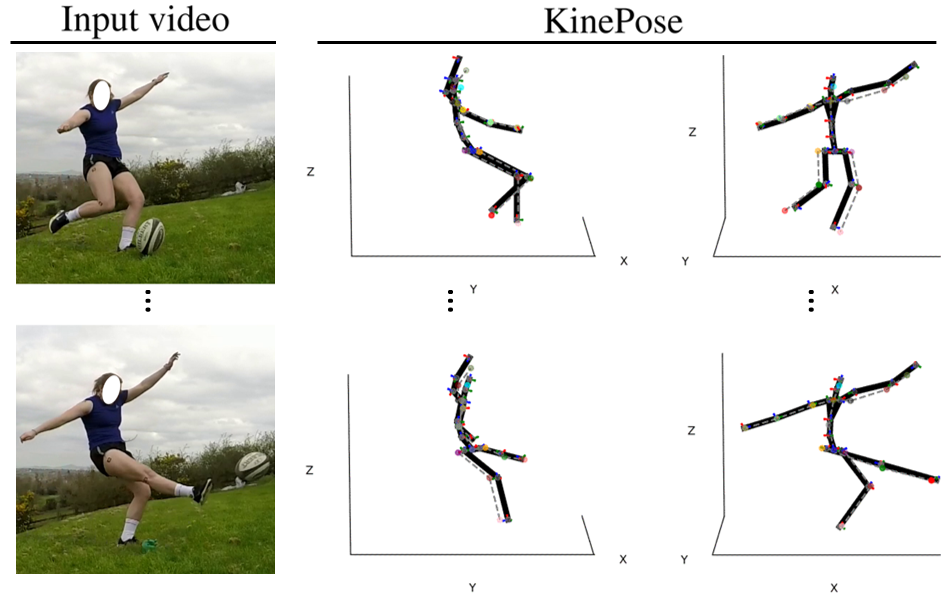}
\end{center}
\vspace{-25pt}
  \caption{KinePose applied to an example case. 3D poses from \cite{Liu2020} (multicolor keypoints and gray dashed lines).}
  \vspace{-10pt}
\end{wrapfigure}

Human motion capture technologies are widely used for animation, physiotherapy, sports biomechanics, ergonomics, robotics, and augmented/virtual reality. However, many of these systems currently involve the use of calibrated multi-sensor setups, making them prohibitively expensive and unsuitable for potential in-the-wild applications. In recent years, a wide variety of computer vision based 3D pose estimation methods have been developed, which present opportunities for a form of free/low-cost markerless motion-capture for use on in-the-wild video footage, e.g., \cite{DineshReddy2021,Pavllo2018a}.  

State-of-the-art position-based 3D pose estimators are capable of inferring root-relative joint positions with remarkable accuracy, however, joint/link orientations are often not included, limiting their practical uses. Approaches that indirectly regress for joint positions through applying joint orientations within a kinematic chain currently have sub-par performance, and do not allow for user definition of chain properties, i.e., limb lengths and joint ranges of motion (ROMs). Therefore, we propose a temporal optimization technique for mapping/retargeting the motion of an ordered set of 3D  joint positions obtained from a pose estimator, to a user-defined open kinematic chain. The kinematic solution space (see \ref{3.2}) is constrained using joint DOFs and ROMs and limiting joint orientation changes between adjacent frames for temporal consistency  (see \ref{3.3}). The proposed approach can be applied as a post-processing technique for position-based 3D pose estimates (particularly useful for monocular pose estimators), allowing for refinement of 3D poses for a subject-specific biomechanical model. 

\vspace{-10pt}
\section{State of the Art}
\vspace{-10pt}
Both multi-camera and single-camera (monocular) 3D human pose estimation methods exist. Generally, they employ a 2D keypoint pose estimator as a backbone and use computer vision/machine learning techniques to ‘lift’ the 2D pose into 3D. A simple calibrated multi-camera technique involves classical algebraic triangulation of keypoint/joint positions. Many machine learning/regression-based multi-camera pose estimators have also been developed, which use neural networks trained and tested on widely used motion capture datasets (e.g. Human3.6M) to regress for both 2D and 3D keypoints directly from images/videos. Regression-based multi-camera pose estimators have achieved high accuracies of below 2cm Mean Per Joint Position Error (MPJPE) over a variety of tasks in Human3.6M \cite{DineshReddy2021,Iskakov2019}. Considering the inherent depth ambiguities, remarkably high accuracy has also been achieved from monocular pose estimators (in the order of 3-5cm MPJPE) \cite{DineshReddy2021,Pavllo2018a}. However, position-based pose representation suffers from several limitations. In particular, joint/link orientations are not included, subject size estimates are inaccurate, and predicted limb lengths are often quite variable, greatly limiting its practical uses, e.g., in biomechanics applications. Other regression-based pose estimators address a portion of these limitations through predictions of joint rotations within a kinematic chain \cite{Pavllo2018,Zhou2016}, i.e., through forward kinematics (FK) (see \ref{3.1}). There are various ways to parameterize these rotations (e.g., sequential Euler/Cardan angles, Euler/Screw axis-angle), though unit quaternions are generally preferred as they overcome singularities associated with operations in $\mathbb{R}^{3}$. Orientation-based 3D pose representation allows for orientation losses to be included, i.e., MPJAS (see \ref{3.3.4}). Furthermore, model-based pose estimators allow for other biomechanical constraints such as joint DOFs. Though orientation-based pose representation is more physically meaningful than position-based representation, the results are mostly sub-par in terms of MPJPE. However, these are promising approaches meriting further development. A popular human biomechanics software OpenSim contains an IK tool for prescribing joint positions to a biomechanical model, which has recently been applied to obtain joint orientations from triangulated 2D predictions, i.e., Pose2Sim \cite{Pagnon2021}. However, many practical applications require a similar tool for 3D pose estimates from a monocular method, which have characteristics that restrict its application. Namely, subject dimensions are inaccurately scaled (since there is no context for the size of the subject), and limb lengths are often variable. Therefore, our approach instead prescribes link directions  (see \ref{3.3.2}), thus allowing for retargeting of incongruent skeletons.

\vspace{-10pt}
\section{Methods}
\vspace{-10pt}
In this section we give an overview of our IK approach for mapping kinematic chains, i.e., kinematic retargeting. In \ref{3.1} we describe the process of FK, which is used in parameterization. In \ref{3.2} we describe the indeterminate nature of the problem. In \ref{3.3}, we describe the proposed IK optimization procedure incorporating chain DOFs, ROMs, and kinematic loss terms.
\vspace{-10pt}
\subsection{Forward kinematics}\label{3.1}
\vspace{-5pt}
Kinematic chains can be described as a hierarchical system of links and joints. They are represented by: 1) the locations and orientations of each of the joints, and 2) the hierarchy of the joints in the system. The Denavit-Hartenberg (DH) FK convention allows for compact representation and convenient manipulation of kinematic chains as a series multiplication of 4x4 transformation matrices along the path to the joint. DH-FK uses a parent-child convention, whereby positions and orientations of child joints are expressed in the parent coordinate systems, i.e., the position of an arbitrary joint ($\mathrm{j}$) is expressed as a vector in the coordinate system of its parent joint/body ($\mathrm{{j}_{parent}}$), i.e., $\mathrm{\left\{\overrightarrow{r}\right\}}_{{j},{j}_{parent}}^{{j}_{parent}}$, and the orientation of the joint is expressed as a 3x3 rotation matrix which specifies the orientation with respect to the parent, i.e., $\left[ \mathrm{R}^{j,{j}_{parent}} \right]$. 
For each joint, the parent-child transformation matrix is then defined in Equation 1. The set of these for an N-joint chain C (with ‘0’ denoting the global coordinate system) is expressed in Equation 2.
\vspace{-10pt}

\noindent
\begin{minipage}{0.49\textwidth}
\begin{align}
\resizebox{.70\hsize}{!}{$
\mathrm{\left\{ \mathrm{T}_{j} \right\}}^{{j},{j}_{parent}}=\left[ \begin{matrix}
\left[ \mathrm{R}^{j,{j}_{parent}} \right] & \mathrm{\left\{\overrightarrow{r}\right\}}_{{j},{j}_{parent}}^{{j}_{parent}} \\
0\ 0\ 0\ & 1
\end{matrix} \right].\;$
}
\end{align}

\end{minipage}
\hfill
\begin{minipage}{0.5\textwidth}
\begin{align}
\resizebox{.50\hsize}{!}{$
\mathrm{C}=\left\{ \mathrm{\left\{ {T}_{1} \right\}}^{1,0}, \dots , \mathrm{\left\{ \mathrm{T}_{N} \right\}}^{N,{N}_{parent}} \right\}.$
}
\end{align}
\end{minipage}

\vspace{3pt}

To perform kinematic operations on an open chain with branching (e.g., the human body) we need also use a directed graph (G) specifying its hierarchical structure. We can then perform FK to express joint/body positions and orientations in the global coordinate system (Equation 3).
\vspace{-5pt}
\begin{align}
\resizebox{.3\hsize}{!}{$
\begin{matrix}
\mathrm{\left\{ \mathrm{T}_{j} \right\}}^{{j},{0}}=\prod_{i=j}^{1}\mathrm{\left\{ \mathrm{T}_{i} \right\}}^{{i},{{i}_{parent}}} \\
\mathrm{for\ i\ on\ path\ to\ root} = \textbf{FK}(C,G).
\end{matrix}$
}
\end{align}

\vspace{-10pt}
The utility of defining a kinematic chain in this manner can be demonstrated if we would like to edit or reorient our kinematic chain. It makes physical sense to apply the rotation in the local coordinate system, e.g., a rotation of the knee is kinematically constrained (due to its DOFs) to occur about a fixed axis defined in the local coordinate system (usually chosen to be a cardinal axis). Furthermore, performing FK with the inclusion of a reorientating transformation matrix $\mathrm{\left\{ \mathrm{T}_{{j}_{\Delta R}} \right\}}^{{j},{{j}_{parent}}}$ will automatically reorient and reposition all down-chain joints (Equation 4). The kinematic chain is parameterized in this way.
\vspace{-5pt}
\begin{align}
\resizebox{.42\hsize}{!}{$
\begin{matrix}
\mathrm{\left\{ \mathrm{T}_{j} \right\}}^{{j},{0}}=\prod_{i=j}^{1}\mathrm{\left\{ \mathrm{T}_{i} \right\}}^{{i},{{i}_{parent}}} \mathrm{\left\{ \mathrm{T}_{{j}_{\Delta R}} \right\}}^{{j},{{j}_{parent}}} \\
\mathrm{for\ i\ on\ path\ to\ root}  = \textbf{FK}(C,G,\mathrm{\left\{ \mathrm{T}_{{j}_{\Delta R}} \right\}}^{{j},{{j}_{parent}}}).
\end{matrix}$
}
\end{align}

\vspace{-10pt}
\subsection{Kinematic indeterminacy}\label{3.2}
\vspace{-5pt}
In the ${\text{Supp.Mat.}}^{\ref{p}}$, we formulate all rotation matrices representing all possible Euler axis-angle combinations for mapping/retargeting a vector $\overrightarrow{a}$ to point in the same direction as a vector $\overrightarrow{b}$ (Equation 5). 
\vspace{-5pt}
\begin{align}
\resizebox{0.95\textwidth}{!}{$
\begin{matrix}
\mathrm{\left[ \mathrm{R}^{\widehat{a}\to \widehat{b}}\right]}^{EAS} =  \\
\left[ \begin{matrix}
{n}_{\alpha,x}^{2} + ({n}_{\alpha,y}^{2}+ {n}_{\alpha,z}^{2})cos({\Phi}_{\alpha}) & {n}_{\alpha,x}{n}_{\alpha,y}(1-cos({\Phi}_{\alpha}))-{n}_{\alpha,z}sin({\Phi}_{\alpha}) & {n}_{\alpha,x}{n}_{\alpha,z}(1-cos({\Phi}_{\alpha}))+{n}_{\alpha,y}sin({\Phi}_{\alpha})\\
{n}_{\alpha,x}{n}_{\alpha,y}(1-cos({\Phi}_{\alpha}))+{n}_{\alpha,z}sin({\Phi}_{\alpha}) & {n}_{\alpha,y}^{2} + ({n}_{\alpha,x}^{2}+ {n}_{\alpha,z}^{2})cos({\Phi}_{\alpha}) & {n}_{\alpha,y}{n}_{\alpha,z}(1-cos({\Phi}_{\alpha}))-{n}_{\alpha,x}sin({\Phi}_{\alpha})\\
{n}_{\alpha,x}{n}_{\alpha,z}(1-cos({\Phi}_{\alpha}))-{n}_{\alpha,y}sin({\Phi}_{\alpha}) & {n}_{\alpha,y}{n}_{\alpha,z}(1-cos({\Phi}_{\alpha}))+{n}_{\alpha,x}sin({\Phi}_{\alpha}) & {n}_{\alpha,z}^{2} + ({n}_{\alpha,x}^{2}+ {n}_{\alpha,y}^{2})cos({\Phi}_{\alpha})
\end{matrix} \right].
\end{matrix}$
}
\end{align}
Where $\mathrm{\overrightarrow{n}}_{\alpha}$ and $\mathrm{\Phi}_{\alpha}$ represent the Euler axis-angle solution space (EAS), and $\alpha$ is a continuous angular value spanning $-\pi$ to $\pi$ used to define unit vectors on the plane symmetrically bisecting the vectors, i.e., candidate Euler axes for which corresponding Euler angles can be computed.

\begin{wrapfigure}{r}{0.6\textwidth}
  \vspace{-10pt}
  \begin{center}
    \includegraphics[width=0.6\textwidth]{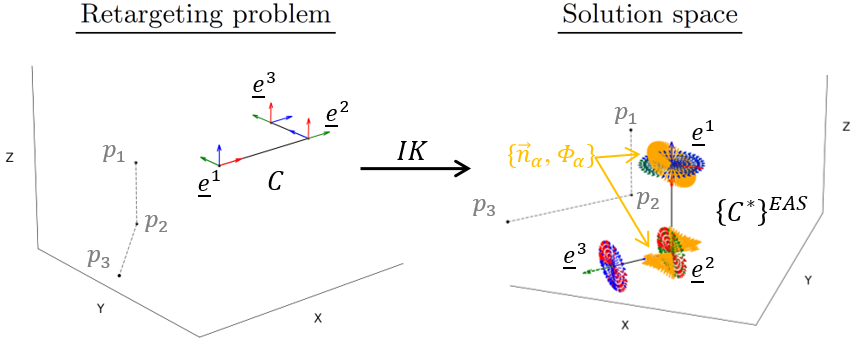}
\end{center}
\vspace{-20pt}
  \caption{Solution space for mapping/retargeting a kinematic chain to an incongruent hierarchical set of 3D positions.}\label{Fig. 2}
  \vspace{-10pt}
\end{wrapfigure}
We can use Equation 5 to analytically define the solution space for retargeting/mapping one kinematic chain (C) to an ordered set of 3D positions (P). The Euler axis-angle solution spaces for all joints in the chain are computed by combining Equations 4\&5. This is visualized in Figure 2 for a 3-joint serial chain. The problem has an indeterminate solution (infinite solution space), therefore, optimization is needed. Specifically, the indeterminacy of this problem is due to the ambiguous ‘twist’ angle. The solution space can be somewhat constrained by specifying anatomical joint DOFs and ROMs (see \ref{3.3.1}). The problem is complicated in a chain such as the human body with branching, i.e., the mapping process used to generate the solution space is faced with a dilemma when reorienting branching joints. Therefore, our approach involves IK optimization, with parameterization of the reorientation transformation matrix in Equation 4 (i.e., $\mathrm{\left\{ \mathrm{T}_{{j}_{\Delta R}} \right\}}^{{j},{{j}_{parent}}}$). In general terms, since motion sequences involve kinematic indeterminacy our hypothesis is that two metrics can be used to converge to an optimal solution: 1) a frame-based pose loss term, and 2) a temporal consistency loss term (see \ref{3.3.2}). 

\vspace{-10pt}
\subsection{Inverse kinematics optimization}\label{3.3}
\vspace{-5pt}
\subsubsection{Kinematic chain}\label{3.3.1}
\vspace{-5pt}

\begin{wrapfigure}{r}{0.35\textwidth}
  \vspace{-50pt}
  \begin{center}
    \includegraphics[width=0.35\textwidth]{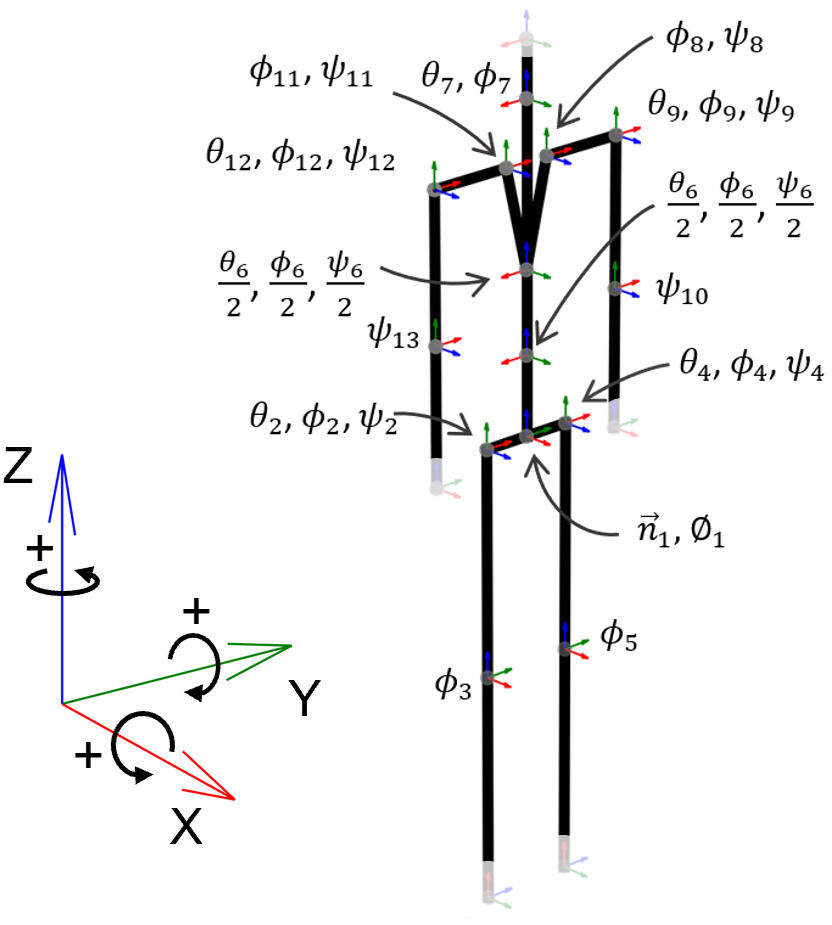}
\end{center}
\vspace{-20pt}
  \caption{Kinematic chain parameters.}\label{Fig. 3}
  \vspace{-10pt}
\end{wrapfigure}
Many position-based 3D pose estimators employ a similar chain configuration. We define a kinematic chain based on this, using common characteristics to optimize for joint orientations. Joints/bodies are defined to correspond to keypoints, i.e., the mid pelvis (defined as the root), hips, knees, ankles, mid spine, neck, head, shoulders, elbows, and ankles. Additional joints are also defined for greater biomechanical fidelity, i.e. the lower spine, and clavicles (see Figure 3). The chain has 14 joints for reorientation, and a total of 28 rotational DOFs. Initially, the kinematic chain is defined in a rest/quiet standing pose, i.e., we define a set of transformation matrices (see Equations 2\&3) to describe a human in a relaxed standing pose with their arms by their sides. Limb/link lengths in the chain are constant and may be user defined. Changes in pose from rest are implemented using FK and an additional transformation matrix (see Equation 4). Joint ROMs may be specified in the local coordinate system as allowable angle ranges about each cardinal axis of the joint coordinate system, where $\mathrm{\theta }_{j}$, $\mathrm{\phi }_{j}$, and $\mathrm{\psi }_{j}$ correspond to sequential rotations about the local X, Y, and Z axes respectively (see Figure 3). For assessment, generalized ROM limits are used, however, they can be further customized. Since there is no 3D keypoint that can be used to directly target the lower spine joint, the lower and mid spine joints are assigned equally proportioned ROMs which add to the overall ROM of that area of the spine. Though not without limitations \cite{Baerlocher2001}, Cardan/Euler angles present a convenient and intuitive approach for parameterizing the optimization weights associated with joint rotations within the kinematic chain, whilst also allowing for specification of joint ROMs using bounds to constrain the solution space. Gimbal lock singularities limit the of use of Cardan/Euler angles for parameterization of human pose. However, the application of anatomical joint DOFs and ROMs in the form of Cardan/Euler angle ranges naturally avoids singularities in our formulation by limiting $-\frac{\pi}{2}<\phi _{j}<\frac{\pi}{2}$ (see ${\text{Supp.Mat.}}^{\ref{p}}$). However, since the root joint retains a full ROM, we parameterize the root orientation with respect to the global coordinate system using the Euler/Screw axis-angle convention ($\overrightarrow{n}=\left\{{n}_{x},{n}_{y},{n}_{z} \right\}$, $\Phi$). Using ROMs throughout the chain and bounding the Screw angle in the root joint also overcomes rotation periodicity, i.e.,
$\sin{x}=\sin{(x+2\pi}$, and $\cos{x}=\cos{x+2\pi}$.

\vspace{-10pt}
\subsubsection{Losses}\label{Losses}\label{3.3.2}
\vspace{-5pt}

\vspace{-0pt}
\begin{wrapfigure}{r}{0.33\textwidth}
  \vspace{-60pt}
  \begin{center}
    \includegraphics[width=0.33\textwidth]{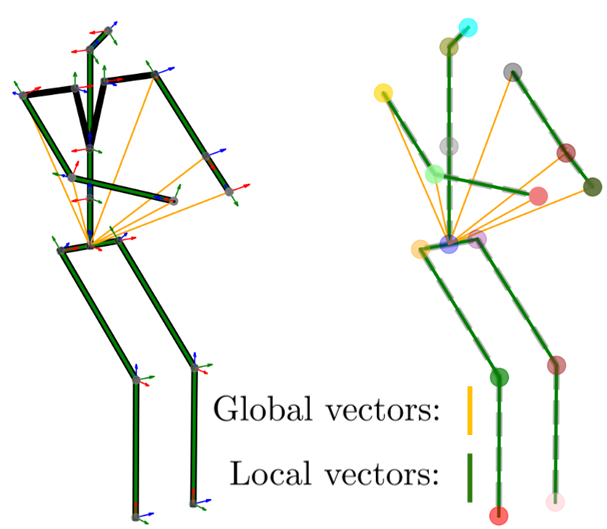}
\end{center}
\vspace{-20pt}
  \caption{Pose loss vectors.}\label{Fig. 4}
  \vspace{-20pt}
\end{wrapfigure}
Two frame-based pose error terms are defined, which both minimize the angular difference in vectors between the 3D pose and the reoriented kinematic chain. We apply a local pose error term (Equation 6) which penalizes angular difference in link vector directions (shown in green in Figure 4), where $\mathrm{\left\{\widehat{r}\right\}}_{{j},{j}_{parent}}^{{0}}$ is the normalized vector in the kinematic chain linking joint j to its parent in the global coordinate system (denoted by superscript ‘0’) and $\mathrm{\left\{\widehat{p}\right\}}_{{j},{j}_{parent}}^{{0}}$ is the corresponding normalized vector on the 3D pose skeleton. Since the kinematic chain includes clavicles which do not correspond to 3D pose keypoints, and there is no appropriate local pose vector to use to constrain the orientations of these joints, we also include a global pose error term (Equation 7) which considers angular difference in global vectors in the arms of the kinematic chain $\mathrm{\left\{\widehat{r}\right\}}_{{j},{1}}^{{0}}$, i.e., stemming from the root joint (shown in orange in Figure 4).
\vspace{-10pt}

\noindent
\begin{minipage}{0.49\textwidth}
\begin{align}
\resizebox{0.90\hsize}{!}{$
\mathrm{E}_{\angle \mathrm{(\overrightarrow{r},\ \overrightarrow{p})}_{local}} = \frac{1}{N}\sum_{j=1}^{N}\left\| \mathrm{cos}^{-1}(\mathrm{\left\{\widehat{r}\right\}}_{{j},{j}_{parent}}^{{0}}\cdot \mathrm{\left\{\widehat{p}\right\}}_{{j},{j}_{parent}}^{{0}}) \right\|.$
}
\end{align}
\end{minipage}
\hfill
\begin{minipage}{0.5\textwidth}
\begin{align}
\resizebox{0.75\hsize}{!}{$
\mathrm{E}_{\angle \mathrm{(\overrightarrow{r},\ \overrightarrow{p})}_{global}} = \frac{1}{A}\sum_{j=1}^{A}\left\| \mathrm{cos}^{-1}(\mathrm{\left\{\widehat{r}\right\}}_{{j},{1}}^{{0}}\cdot \mathrm{\left\{\widehat{p}\right\}}_{{j},{1}}^{{0}}) \right\|.$
}
\end{align}
\end{minipage}
\vspace{5pt}

In cases where there are co-linear sequential links, i.e., when there are fully extended limbs (knees or elbows), there will be kinematic indeterminacy. Our hypothesis is that appropriately weighted minimization of joint orientation changes between frames ($\mathrm \Delta \mathrm{\Theta}_{j}$) can constrain the solution space toward an optimal solution with temporal consistency, i.e., optimizing poses across frames with extended and bent limbs (Equation 8) (for a sequence of poses M frames long).
\vspace{-5pt}
\begin{align}
\resizebox{0.3\hsize}{!}{$
\mathrm{E}_{\Delta \mathrm{\Theta}} = \frac{1}{M}\sum_{m=1}^{M}  \left\{ \sum_{j=1}^{N} \left\| \mathrm \Delta \mathrm{\Theta}_{j} \right\| \right\}_{m},$
}
\end{align}
\vspace{-10pt}

where $\mathrm \Delta \mathrm{\Theta}_{j} = \mathrm{\Theta}_{j}^{m+1}-\mathrm{\Theta}_{j}^{m}$ for $m=1$, $\mathrm \Delta \mathrm{\Theta}_{j} = (\mathrm{\Theta}_{j}^{m+1}-\mathrm{\Theta}_{j}^{m-1})/2$ for $1 < m < M$, and $\mathrm \Delta \mathrm{\Theta}_{j} = \mathrm{\Theta}_{j}^{m}-\mathrm{\Theta}_{j}^{m-1}$ for $m=M$.

We define a loss function $\mathrm L_{frame}$ for frame-by-frame IK (Equation 9), and combining Equations 6,7\&8 with scalar weighting for the temporal term which competes with the pose error terms, we define $\mathrm L_{temporal}$ for temporal IK (Equation 10).

\vspace{-10pt}
\noindent
\begin{minipage}{0.49\textwidth}
\begin{align}
\resizebox{0.7\hsize}{!}{$
\mathrm L_{frame} = \mathrm{E}_{\angle \mathrm{(\overrightarrow{r},\ \overrightarrow{p})}_{local}} + \mathrm{E}_{\angle \mathrm{(\overrightarrow{r},\ \overrightarrow{p})}_{global}},$
}
\end{align}
\end{minipage}
\hfill
\begin{minipage}{0.5\textwidth}
\begin{align}
\resizebox{0.72\hsize}{!}{$
\mathrm L_{temporal} = \frac{1}{M}\sum_{m=1}^{M}\mathrm L_{frame=m} +\lambda \mathrm{E}_{\Delta \mathrm{\Theta}}.$
}
\end{align}
\end{minipage}

\vspace{-10pt}
\subsubsection{Algorithms}\label{Algorithms}
\vspace{-5pt}
A sequential least squares programming procedure is used to solve the minimization problem \cite{Kraft1988}. This procedure is commonly used for robust nonlinear programming solutions to kinematic retargeting and IK. All algorithms take the kinematic chain C in its rest/quiet pose, and a sequence of 3D poses for F frames as inputs (P). The algorithms output optimized model parameters or joint DOFs ($\mathrm{\Theta}^{*}$) using IK (Equation 11). These can then be used to specify a sequence of retargeted chains $ \left\{ \mathrm{C}_{1}^{*} , \dots , \mathrm{C}_{F}^{*} \right\} $ using Equation 4.
\vspace{-5pt}
\begin{align}
\resizebox{0.45\hsize}{!}{$
\mathrm{\Theta}^{*}=\textbf{IK}(C,G,P,\mathrm{\Theta}_{0})\ \mathrm{subject\ to\ }  \arg \min_{\Theta} (\mathrm L).$
}
\end{align}
\vspace{-20pt}

Since convergence of local optimization methods depends on the initial guess ($\mathrm{\Theta}_{0}$), we perform a relatively computationally inexpensive frame-by-frame pre-optimization as input to our temporal optimization, using $\mathrm{L}_{frame}$  (Equation 9) as the objective function (Algorithm 1). This addition also speeds up the optimization time significantly.
\begin{wrapfigure}{r}{0.75\textwidth}
  \vspace{-30pt}
  \begin{center}
    \includegraphics[width=0.75\textwidth]{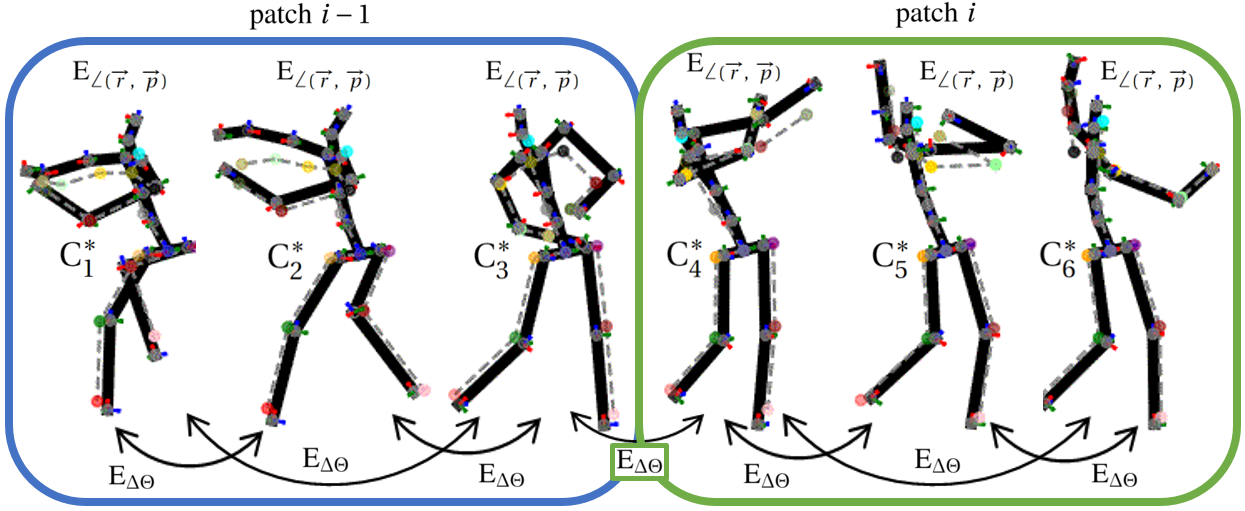}
\end{center}
\vspace{-25pt}
  \caption{Overview of the proposed IK approach (patch length M=3).}\label{Fig. 5}
  \vspace{-15pt}
\end{wrapfigure}

The temporal algorithm (Algorithm 2) involves blocks, or patches of M frames (out of a total of F frames) in which sequences of poses are retargeted in tandem with the inclusion of both pose and temporal error terms (Equation 10). Patches are temporally linked or ‘stitched’ with an additional temporal error term, i.e., the difference in model parameters ($\mathrm{E}_{\Delta \mathrm{\Theta}}$) between the final frame of patch $i-1$ and the first frame of patch $i$ are used in the loss function for patch $i$ (Figure 5).

  \begin{center}
    \includegraphics[width=1\textwidth]{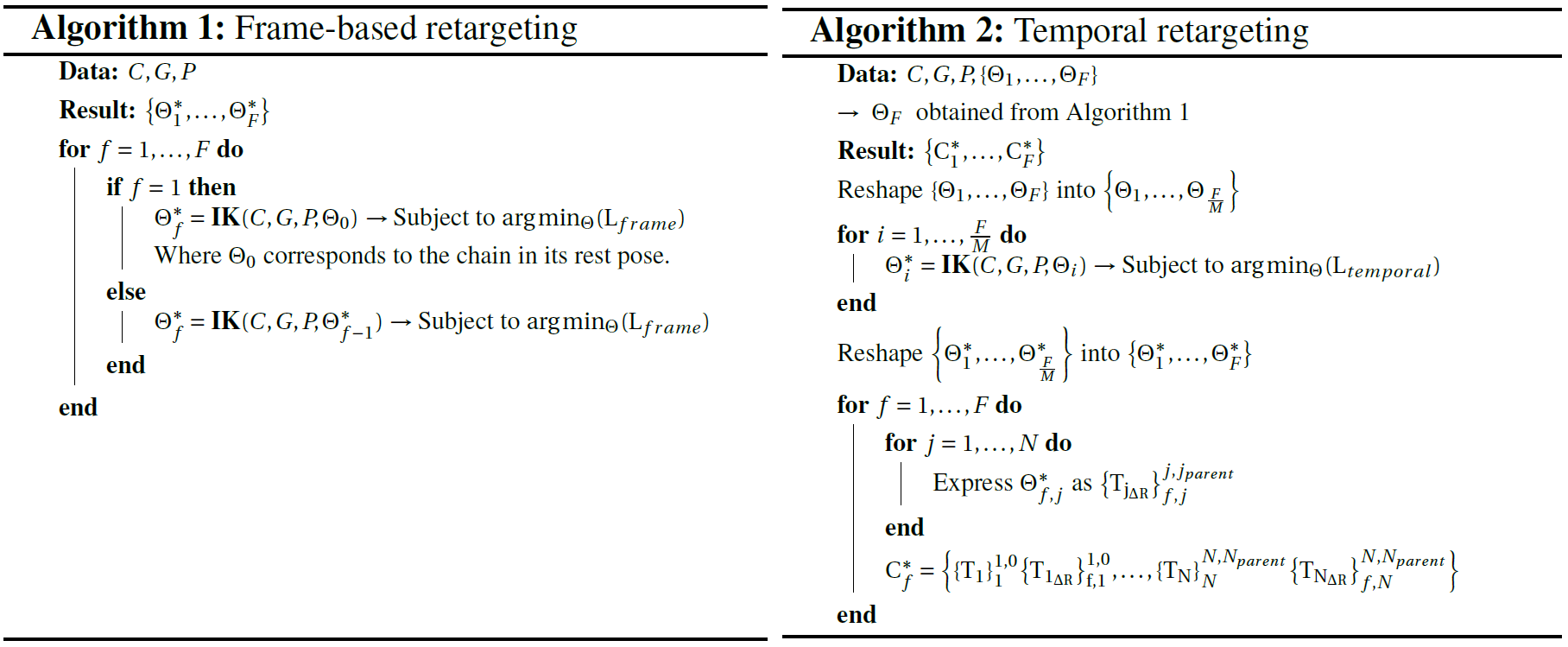}
\end{center}
\vspace{-20pt}

\subsubsection{Assessment}\label{3.3.4}
\vspace{-5pt}
For algorithmic assessment of the IK approaches, we generate twenty-four 30-frame long motion sequences within joint ROMs for the kinematic chain, with three levels of motion speed, with maximum cardan angle changes between frames of a) $\frac{\pi}{500}$, b) $\frac{\pi}{200}$, and c) $\frac{\pi}{70}$. The latter corresponding approximately to full movement within ROMs. Half of the sequences consist of only bent limbs, and the other half have a combination of prescribed phases of either bent or extended limbs. The motion sequences have been defined such that each sequence with bent limbs has a corresponding sequence with phases of bent limbs and extended limbs, where motion in the rest of the skeleton is identical. These phases are: 1) 3 frames extended, 2) 9 frames bent, 3) 6 frames extended, 4) 9 frames bent, 5) 3 frames extended. Generation of motion in this manner allows for analysis of the effect on accuracy of extended limbs (which results in ambiguity in twist angle) for various IK implementations.
We apply a series of IK algorithms to the resulting joint pose sequences, i.e., with ground truth joint orientations hidden. We compare Algorithm 1 with and without feeding of optimized model parameters from previous frames ($\mathrm{1}_{a}$ and $\mathrm{1}_{b}$ respectively), and Algorithm 2 with different patch sizes, i.e., $\mathrm{2}_{3}$ (M=3) and $\mathrm{2}_{5}$ (M=5). Computation time on an Intel\textsuperscript{\textregistered} Core\textsuperscript{\texttrademark} i7-9700 processor is recorded for each implementation. Overall agreement was assessed using Mean Per Joint Angular Separation ($\mathrm{MPJAS}_{N}$) for all joints that are reoriented (N=14) across frames, i.e., the average angular difference about the Screw axis between inferred and ground truth (GT) joint orientations $\mathrm{\left\{ \mathrm{\Phi}_{j} \right\}}_{pred,GT}$ (Equation 12). For comparison, we also calculate $\mathrm{MPJAS}$ for each individual joint.
\vspace{-20pt}

\begin{align}
\resizebox{0.39\hsize}{!}{$
\mathrm{MPJAS}_{N}=\frac{1}{F}\sum_{f=1}^{F}   \left\{ \frac{1}{N}\sum_{j=i}^{N}\left\| \mathrm{\left\{ \mathrm{\Phi}_{j} \right\}}_{pred,GT} \right\| \right\}_{f}.$
}
\end{align}

An ablation study was performed on a sample of 12 additional motion sequences to determine the optimal scalar weight for the temporal model ($\lambda$) for different motion speeds ($\lambda = 0.3$ for speed c, $\lambda = 0.5$ for speed b, and $\lambda =  0.7$ for speed a).

\vspace{-15pt}
\section{Results}
\vspace{-10pt}
Table 1 shows a comparison of average angular errors and optimization times for Algorithm 1 with feeding of weights between frames for initialization ($\mathrm{1}_{b}$), or no feeding ($\mathrm{1}_{a}$), and Algorithm 2 with patches of length 3 ($\mathrm{2}_{3}$) or 5 ($\mathrm{2}_{5}$) for three levels of motion speed.
\vspace{-15pt}
\begin{table}[H]
\begin{center}
\caption{Accuracy of IK Algorithms 1 and 2. $\mathrm{MPJAS}_{14}$ per frame for three levels of motion speed (fps: Optimization frames per second, E: 14-joint Mean Per Joint Angular Separation (rad/joint)).}\label{Table 1}
\vspace{-10pt}
\resizebox{\textwidth}{!}{%
\begin{tabular}{clccccccccccc}
\multicolumn{1}{l}{} &  & \multicolumn{2}{c}{\textbf{Speed a}} &  & \multicolumn{2}{c}{\textbf{Speed b}} &  & \multicolumn{2}{c}{\textbf{Speed c}} &  & \multicolumn{2}{c}{\textbf{Average}} \\ \cline{3-4} \cline{6-7} \cline{9-10} \cline{12-13} 
\textbf{Algorithm}   &  & fps              & E                 &  & fps              & E                 &  & fps              & E                 &  & fps             & E                \\ \cline{1-1} \cline{3-4} \cline{6-7} \cline{9-10} \cline{12-13} 
$\mathrm{1}_{a}$                   &  & 7.44x10$^{-2}$        & 9.45 x10$^{-2}$        &  & 6.96x10$^{-2}$        & 9.13 x10$^{-2}$        &  & 7.48x10$^{-2}$        & 8.12x10$^{-2}$        &  & 7.29x10$^{-2}$       & 8.90 x10$^{-2}$       \\
$\mathrm{1}_{b}$                   &  & 9.63x10$^{-2}$        & 7.65x10$^{-2}$         &  & 9.28x10$^{-2}$        & 7.45x10$^{-2}$        &  & 8.61x10$^{-2}$        & 5.68x10$^{-2}$        &  & 9.15x10$^{-2}$       & 6.93x10$^{-2}$       \\
$\mathrm{2}_{3}$                   &  & 2.03x10$^{-2}$        & 8.28x10$^{-2}$        &  & 2.33x10$^{-2}$        & 7.69x10$^{-2}$        &  & 5.74x10$^{-2}$        & 5.07x10$^{-2}$        &  & 2.74x10$^{-2}$       & 7.01x10$^{-2}$       \\
$\mathrm{2}_{5}$                   &  & 8.94x10$^{-3}$        & 7.07x10$^{-2}$        &  & 1.32x10$^{-3}$        & 7.40x10$^{-2}$        &  & 3.90x10$^{-2}$        & 5.13x10$^{-2}$        &  & 1.41x10$^{-2}$       & 6.53x10$^{-2}$      
\end{tabular}
}
\end{center}
\end{table}
\vspace{-25pt}

\begin{wraptable}{r}{9cm}
\vspace{-25pt}
\begin{center}
\caption{$\mathrm{MPJAS}_{14}$ per frame for motion sequences with 1) only bent limbs, or 2) phases of extended limbs.}\label{Table 2}
\resizebox{0.5\textwidth}{!}{%
\begin{tabular}{ccccc}
\textbf{Algorithm} &  & \textbf{Bent limbs only} &  & \textbf{Extended \& bent limbs} \\ \cline{1-1} \cline{3-3} \cline{5-5} 
$\mathrm{1}_{a}$                 &  & 5.84x10$^{-2}$               &  & 1.20x10$^{-1}$                      \\
$\mathrm{1}_{b}$                 &  & 6.05x10$^{-2}$               &  & 7.80x10$^{-2}$                      \\
$\mathrm{2}_{3}$                 &  & 5.39x10$^{-2}$               &  & 8.64x10$^{-2}$                      \\
$\mathrm{2}_{5}$                 &  & 5.64x10$^{-2}$               &  & 7.43x10$^{-2}$        
\end{tabular}
}
\end{center}
\vspace{-25pt}
\end{wraptable}
Table 2 shows a comparison of average angular errors for motions with 1) bent limbs only, or 2) a combination of bent and extended limbs. Motion sequences with only bent limbs have low errors for all algorithms, and for sequences with phases of extended limbs algorithm $\mathrm{1}_{b}$ and both temporal algorithms have lower errors.

Figure 6 shows a comparison of average average angular errors for parameterized joints in the chain for motion sequences with phases of bent and extended limbs. Algorithm $\mathrm{1}_{a}$ produces higher errors in the clavicles (due to a lack of corresponding keypoints), and shoulders and hips (associated with ambiguity in twist angles). Results for motion sequences with only bent limbs are provided in the ${\text{Supp.Mat.}}^{\ref{p}}$ Furthermore, the relationship between phases of extended/bent limbs are investigated further for boundary cases.
\vspace{-10pt}
\begin{figure}[H]
\centering
\includegraphics[width=17cm]{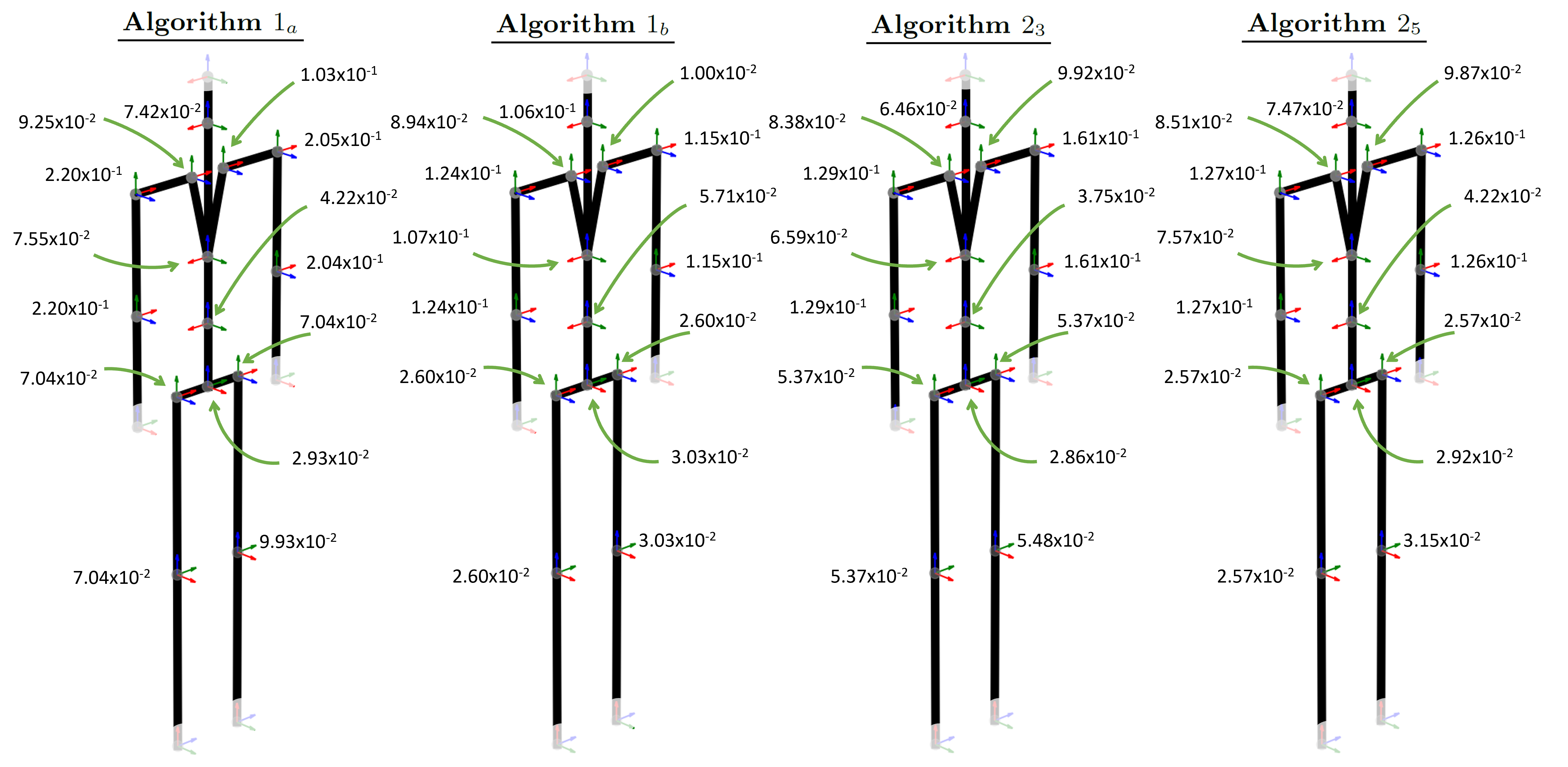}
\vspace{-30pt}
\caption{$\mathrm{MPJAS}_{1}$ by joint and algorithm, for motion sequences with phases of bent and extended limbs.}
\label{fig:example}
\end{figure}

\vspace{-30pt}

\section{Discussion}
\vspace{-10pt}
$\mathrm{MPJAS}_{14}$ errors are low for all algorithms and motion speeds, ranging from average errors of 8.90x10$^{-2}$ rad/joint for algorithm $\mathrm{1}_{a}$ (5.1\textdegree), to 6.53x10$^{-2}$ rad/joint for algorithm $\mathrm{2}_{5}$ (3.7\textdegree) (Table 1). Differences observed between algorithm $\mathrm{1}_{a}$ and $\mathrm{1}_{b}$ highlight the importance of the initialization guess for weights, i.e., feeding weights from previous frames for initialization in a frame-by-frame optimization reduces time to convergence (Table 1), and adds a degree of temporal consistency. The results provide a clear justification for the temporal model. While algorithm ${1}_{b}$ outperforms ${2}_{3}$ on average, temporal algorithm ${2}_{5}$ is the best performer across all indicators. The benefits of the temporal model relate to twist angle ambiguity (Table 2 $\&$ Figure 6), and are of particular benefit for a specific boundary case involving an initial phase of extended limbs, followed by a phase of bent limbs (see ${\text{Supp.Mat.}}^{\ref{p}}$). Large fluctuations in twist angles are observed for algorithm $\mathrm{1}_{a}$, whereas, algorithm $\mathrm{1}_{b}$ and both variations of the temporal algorithm result in lower errors both overall, and in boundary cases. Algorithm $\mathrm{2}_{5}$ achieves the highest accuracy for all motion speeds, and obtains particularly low errors for the lower limbs (hips $\&$ knees), i.e., 2.71x10$^{-2}$ rad/joint (1.6\textdegree) for motion sequences involving phases of extended limbs (Figure 6), and 1.63x10$^{-3}$ rad/joint (0.1\textdegree) for motion with only bent limbs (see ${\text{Supp.Mat.}}^{\ref{p}}$). Temporal implementations with sufficiently large patch sizes are effective for reducing ambiguity in twist angle, which is a key benefit. Nevertheless, algorithm $\mathrm{1}_{b}$ is suitable for applications requiring faster processing. The temporal scalar weight should be tuned for both motion speed and video framerate. For fast motion or a low framerate, relative joint orientation changes between frames will naturally be higher, therefore requiring a lower weighting. Similarly, for slow motion or high framerate a higher weighting is required. 

Due to a lack of relevant keypoints, twist angles of the forearms and neck, and joint reorientations of the ankles and wrists are currently not included. However, an extension could include these for 3D pose inputs with more keypoints in the extremities. A large proportion of the errors are associated with those joints in the chain without corresponding keypoints, i.e., clavicles and lower spine. Parameterization of the lower/mid spine with equally proportioned rotations in two joints may often not be biofidelic, indeed, this relationship is often the subject of analyses in biomechanics. These limitations are imposed by the fact that current 3D pose estimators do not include sufficient keypoints. Synthetic generation of motion, as per \cite{Beeson2015}, allows for a detailed assessment of the mathematical procedure including boundary cases of interest. However, for potential practical applications further testing should include GT activity-specific motion capture footage. Furthermore, practical applications rely on the accuracy of the 3D pose estimator used; calibrated multi-camera pose estimators have higher accuracy, whereas monocular pose estimators are more practically useful. KinePose is applicable to either multi-camera, or monocular 3D pose estimators with incorrectly scaled skeletons and variable link lengths.
\vspace{-15pt}

\section{Conclusions}
\vspace{-10pt}
This paper proposes KinePose, a convenient post-processing technique for position-based 3D human pose estimation methods, to infer joint orientations in a subject-specific biomechanical model. We demonstrate that this technique allows for accurate prediction of joint orientations. This technique may be particularly useful for sports biomechanics, and telehealth applications. Though currently unsuited to real-time applications, we envisage many potential applications as a post-processing technique for 3D pose estimators.

\vspace{-10pt}

\section*{Acknowledgements}

\vspace{-10pt}
This research was funded under the RSA-Helena Winters Scholarship for Studies in Road Safety.
\vspace{-10pt}
%%%%%%%%%%%%%%%%%%%%%%%%

\bibliographystyle{apalike}

\begin{small}
\bibliography{main}
\vspace{-10pt}
\end{small}

\end{document}